\pdfoutput=1

\documentclass[11pt]{article}

\usepackage[]{acl}

\usepackage{times}
\usepackage{latexsym}

\usepackage{adjustbox}
\usepackage{graphicx}
\usepackage{multirow}
\usepackage{amsmath}  
\usepackage{amsfonts} 
\usepackage{url}
\usepackage{subcaption}

\usepackage{arydshln}
\usepackage{booktabs} 
\usepackage{makecell}
\usepackage{pifont}
\usepackage{algorithm}
\usepackage{algorithmic}

\usepackage[T1]{fontenc}
\usepackage[utf8]{inputenc}
\usepackage{microtype}
\usepackage{inconsolata}

\newcommand{\sysname}{Unified Layer Skipping}
\newcommand{\sysnameshort}{Unified Skipping}

\title{Accelerating Inference in Large Language Models with a \\Unified Layer Skipping Strategy}

\author{
  Yijin Liu,~
  Fandong Meng\thanks{\quad  Corresponding author.}~
  and Jie Zhou \\
  Pattern Recognition Center, WeChat AI, Tencent Inc, China \\
  \texttt{\{yijinliu, fandongmeng, withtomzhou\}@tencent.com} \\
}

\begin{document}
\maketitle

\begin{abstract}
Recently, dynamic computation methods have shown notable acceleration for Large Language Models (LLMs) by skipping several layers of computations through elaborate heuristics or additional predictors. However, in the decoding process of existing approaches, different samples are assigned different computational budgets, which cannot guarantee a stable and precise acceleration effect. Furthermore, existing approaches generally skip multiple contiguous layers at the bottom or top of the layers, leading to a drastic change in the model’s layer-wise representations, and thus a consequent performance degeneration. Therefore, we propose a \sysname~strategy, which selects the number of layers to skip computation based solely on the target speedup ratio, and then skips the corresponding number of intermediate layer computations in a balanced manner. Since the \sysname~strategy is independent of input samples, it naturally supports popular acceleration techniques such as batch decoding and KV caching, thus demonstrating more practicality for real-world applications. Experimental results on two common tasks, {\em i.e.,} machine translation and text summarization, indicate that given a target speedup ratio, the \sysname~strategy significantly enhances both the inference performance and the actual model throughput over existing dynamic approaches.

\end{abstract}

\section{Introduction}
Large-scale pre-trained Language Models (LLMs) have experienced rapid development in recent years and have demonstrated promising performance across various domains~\cite{chatgpt1,chatgpt2,chatgpt3,chatgpt4,chatgpt5,chatgpt6,instructgpt_2022}. However, due to their autoregressive prediction nature, the inference process of LLMs is often computationally expensive. 
Various approaches have been proposed to accelerate the inference speed of LLMs, such as model quantization~\cite{quant_2019,llm_quant_2022,llm_quant_2023,llm_quant1_2023,llm_quant2_2023}, knowledge distillation from larger-scale models~\cite{llm_kd_2021,llm_kd_2023,llm_kd1_2023}, and model pruning~\cite{layerdrop_2019,llm_prune1_2023,llm_prune_2023}. 

There is also a category of dynamic computation methods that do not require changes to the parameters or structure of the model as in the above methods, and also have some complementarity with the above methods.
Specifically, they adaptively allocate different computational budgets based on the specific characteristics of the input samples, thereby achieving overall acceleration without compromising overall performance.
These methods are commonly referred as "early exit" strategies~\cite{ee_2016,ee_2017,ee_2021,da_transformer_2019,calm_2022}. Generally, existing methods design input-aware metrics to determine how many layers of the current sample should be executed, including model intrinsic prediction confidence, additional classifiers that decide whether to exit the computation early, or predefined computational strategies.

Despite the effectiveness of these dynamic computation methods, they present several limitations. First, in the decoding process, different samples are assigned different computational budgets, which cannot guarantee a stable and precise acceleration effect. This variability in computational allocation can lead to unpredictable inference times, which is not ideal for real-world applications where consistent response times are often required.
Second, existing approaches generally skip multiple contiguous layers at the bottom or top of the layers. This approach can lead to a drastic change in the model’s layer-wise representations, and thus a consequent degeneration in performance. The layer-wise representations in LLMs are crucial as they capture different levels of semantic information, and skipping contiguous layers could result in the loss of important information, thereby affecting the model's performance.

To address these issues, we propose a novel strategy, which we refer as \sysname. Unlike existing methods, the \sysname~strategy selects the number of layers to skip computation based solely on the target speedup ratio, and then skips the corresponding number of intermediate layer computations in a balanced manner. This approach ensures a more stable and predictable acceleration effect, as the computational budget is consistent across different samples.
Furthermore, by skipping layers in a balanced manner across the layers, the \sysname~strategy minimizes the changes on the model's layer-wise representations, thereby mitigating the performance degradation observed in existing methods.
Importantly, the \sysname~strategy is independent of the input sample, which means it naturally supports popular acceleration techniques such as batch decoding and KV caching. 

We evaluate the \sysname~strategy on two common tasks, i.e., machine translation and text summarization. Our experimental results indicate that given a target speedup ratio, the \sysname~strategy significantly enhances the inference performance and the actual model throughput over existing dynamic approaches. 
For example, \sysname~achieve about 30\% to 70\% throughput improvements are observed when compared to the existing methods, while ensuring the minimum performance loss at the same speedup effect.
These findings highlight the potential of the \sysname~strategy as a practical and effective solution for accelerating the inference process of LLMs.
Our contribution can be summarized as follows\footnote{Codes are released at \url{https://github.com/Adaxry/Unified_Layer_Skipping}.}:

\begin{itemize}

\item We propose a novel dynamic computation strategy, \sysname, which determines the number of layers to skip based solely on the target speedup ratio. This approach ensures a stable and predictable acceleration effect, as the computational budget is consistent across different samples.

\item Unlike existing methods that skip multiple contiguous layers, the \sysname~strategy skips the corresponding number of intermediate layer computations in a balanced manner. This approach minimizes the impact on the model's layer-wise representations, thereby mitigating the performance degradation observed in existing methods.

\item The \sysname~strategy is independent of the input sample, which makes it naturally compatible with popular acceleration techniques such as batch decoding and KV caching. This feature makes the \sysname~strategy more practical for real-world applications.

\item We provide extensive experimental evidence on two common tasks, i.e., machine translation and text summarization. Our results demonstrate that, given a target speedup ratio, the \sysname~strategy significantly enhances the inference performance and the actual model throughput over existing dynamic approaches. 

\end{itemize}

\begin{figure*}[t!]
\begin{center}
     \scalebox{0.9}{           \includegraphics[width=1\textwidth]{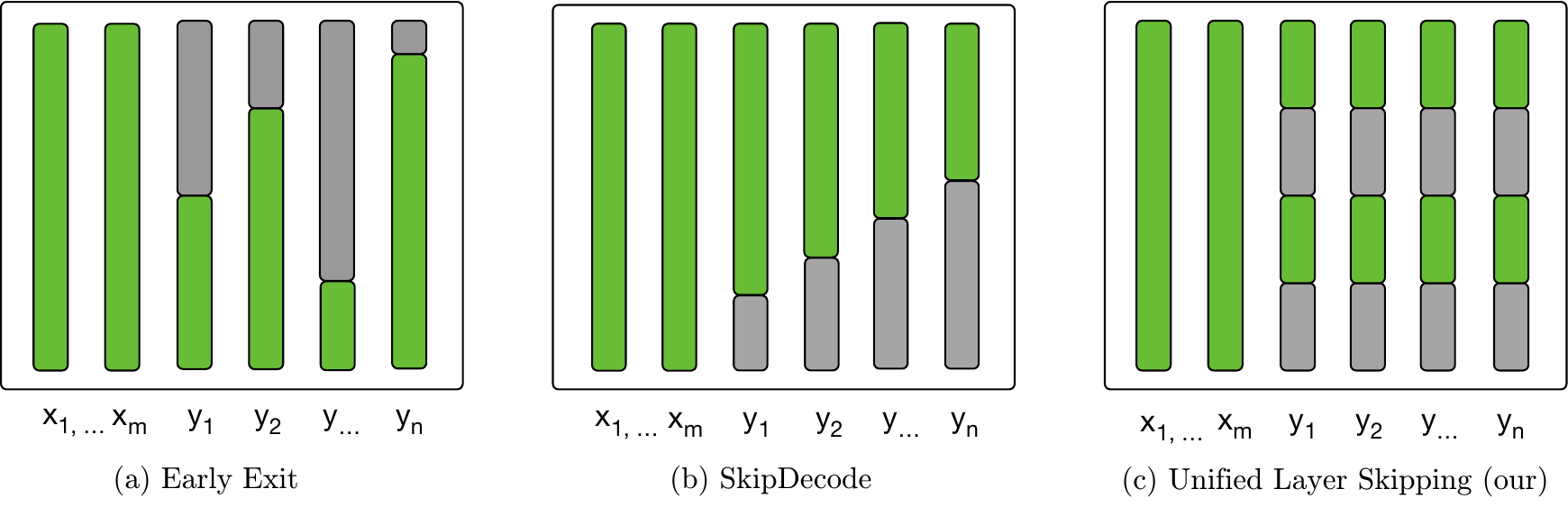}
      } 
      \caption{
       Overview comparisions of serveral related approaches. The gray columns represent the layers that are skipped out, and the green columns represent the layers that are actually activated. $\{x_1, \dots, x_{m}\}$ are the prompt input tokens and $\{y_1, \dots, y_{n}\}$ are the response tokens.
      } 
      \label{fig:overview}  
 \end{center}
\end{figure*}

\section{Related Works}

\paragraph{Instruction Fine-Tuning for LLMs.}
To ensure that the outputs of language models align with human intentions and to harness their full capabilities, InstructGPT~\cite{instructgpt_2022} introduces a method for fine-tuning Large Language Models (LLMs) using a limited set of instruction-following data to align the model's behavior with human preferences. This approach to instruction following has garnered significant interest in both the academic and industrial sectors~\cite{follow_instructgpt1,follow_instructgpt2,follow_instructgpt3,follow_instructgpt4,follow_instructgpt5,follow_instructgpt6}.
Typically, the instruction-following data is composed of two key components, namely, the prompt input tokens $\{x_1, \dots, x_{m}\}$ and the response tokens $\{y_1, \dots, y_{n}\}$. 
During instruction fine-tuning, the cross-entropy loss $L_{CE}$ is generally used for model training, which measures the difference between the predicted probability distribution over the response tokens and the true distribution:
\begin{equation}
L_{CE} = -\sum_{i=1}^{n} \log(p(y_i | x_1, \dots, x_m))
\end{equation}
where $p(y_i | x_1, \dots, x_m)$ represents the model's predicted probability of the $i$-th response token given the prompt input tokens.

\paragraph{Early Exit.}
The Early Exit method is initially introduced to enhance the efficiency of neural networks by ~\citet{ACT_2016}. As shown in Figure~\ref{fig:overview} (a), it incorporates a halting unit that evaluates the continuation or termination of computation at each word in a sentence, thereby determining the probability of proceeding or halting layer-by-layer. 
Early Exit has been successfully adapted to various models, including ResNet~\cite{ACT_resnet_2017}, the standard Transformer architecture~\cite{da_transformer_2019}, and the Universal Transformer~\cite{UT_2019}.
\begin{table*}[t]
\begin{center}
\scalebox{0.8}{
\begin{tabular}{lccccc}
\toprule
\multirow{2}{*}{\textbf{\makecell{Approach}}} &  \multirow{2}{*}{\textbf{\makecell{Batch Decoding}}} & \multirow{2}{*}{\textbf{\makecell{KV Caching}}} &  \multirow{2}{*}{\textbf{\makecell{Stable Speedup}}} & \multirow{2}{*}{\textbf{\makecell{Banlanced Sub-Network}}}  \\
~ & ~ & ~ & ~    \\
\midrule
Early Exit  & $\times$ & $\times$ & $\times$ & $\times$ \\
SkipDecode & $\checkmark$ & $\checkmark$  & $\times$ & $\times$ \\
\sysname~(ours) & $\checkmark$ & $\checkmark$ & $\checkmark$  & $\checkmark$  \\
\bottomrule
\end{tabular}
}
\caption{Comparisons of serveral related approaches. The Early Exit methods are incompatible with common acceleration techniques, {\em e.g.,} batch decoding and KV caching, and do not guarantee a stable acceleration. 
Similarly, SkipDecode also lacks stable and concrete control over the speedup ratio and results in an imbalanced subnetwork structure after layer skipping. 
In contrast, our \sysname~supports all of the aforementioned features.
}
\label{tab:comparision}
\end{center}
\end{table*}
The core principle of Early Exit is to dynamically allocate varying computational costs to different input samples, with the primary distinction lying in the method of allocation. Existing works mainly differ in their strategies for computation allocations, such as utilizing internal model predictions of confidence~\cite{da_transformer_2019,lite_2023} or pre-determined strategies~\cite{faster_da_2021,skipdecode_2023} based on mutual information.
Recently, ~\citet{calm_2022} introduce a dynamic computation method tailored for decoder-only structures, termed Confident Adaptive Language Modeling (CALM). Subsequently, ~\citet{lite_2023} propose to leverage the confidence of intermediate layer predictions to more stably control the speedup.

\paragraph{Skip Decoding.}
Although Early Exit and its varients achieve some success for model acceleration, they prevent the model from using the standard acceleration strategies commonly used during decoding, namely batch decoding and key-value (KV) caching. This is because the model needs to wait for the last token to exit the computation within the batch, and an inconsistent caching size due to dynamic calculations.
Skip Decoding~\cite{skipdecode_2023} is further proposed to pre-define position-wise exit point for every token at a given sequence position.
As shown in Figure~\ref{fig:overview} (b), Skip Decoding requires a linear monotonic skipping of layer computations as the sequence length increases. 
This decouples the skipping strategy from specific input samples and associates it solely with sequence positions, enabling Skip Decoding to be compatible with commonly used modern acceleration techniques, namely batching decoding and kv caching. However, during actual decoding, the model cannot predict the length of the decoding in advance, which may result in an unstable and uncontrollable acceleration ratio.

\section{Approach}
In this section, we initially address the challenges faced by existing input-aware dynamic computation methods in practical applications, namely the issue of unstable acceleration and the unbalanced sub-network structures. Subsequently, we introduce the proposed \sysname~strategy.

\subsection{Challenges in Input-Aware Dynamic Computation Strategies}
The Early Exit series of methods employs various heuristic rules or an additional layer-wise classifier to determine the depth of computation required for different tokens within an input sample. While this input-aware strategy is intuitive, it limits its practical application value. Specifically, with the commonly used batch decoding techniques, Early Exit cannot ensure that different samples within a batch have the same exit points or activate the same sub-network structures, thus preventing the support for batch decoding.
Similarly, for the KV caching technique that often used in autoregressive prediction models, if the depth of computation for a subsequent token exceeds that of a cached token, the current cache contents need to be recalculated, leading to computational inefficiency and waste.

To mitigate these issues, Skip Decoding proposes predefining exit points for all samples according to their sequential lengths and ensuring that the depth executed by subsequent tokens is always less than that executed by earlier tokens.
Although this carefully designed strategy can mitigate the aforementioned computational issues to some extent, it still has several limitations. 
Firstly, in actual decoding processes, 
due to the cumulative error effect, the prediction difficulty of later positions is often greater than that of earlier positions. However, Skip Decoding allocates less prediction cost to these more difficult token predictions, which is inconsistent with the observations and conclusions of existing studies~\cite{accum_error_2020,ss_decoding_2021}. 
Secondly, the final decoding length cannot be known in advance during actual decoding, so in most cases, the model will still execute more depth at relatively earlier positions according to the above strategy, resulting in a limited and unstable acceleration ratio. 
Ultimately, both Early Exit and Skip Decoding continuously skip the computation of several top or bottom layers of the model, leading to an imbalance in the sub-network structure after skipping layers, which in turn results in more drastic changes in the model's layer-wise representations and a more pronounced degradation in performance.

\begin{table*}[t]
\begin{center}
\scalebox{0.7}{
\begin{tabular}{ccll}
\toprule
\multirow{2}{*}{\textbf{\makecell{Target\\Speedup}}} &  \multirow{2}{*}{\textbf{\makecell{\#Target Activated \\ Layer}}}  & \multirow{2}{*}{\textbf{Skipping Strategies}} &  \multirow{2}{*}{\textbf{\makecell{Indexs of Layers to Retain after Layer Skipping }}} \\
~ & ~   \\
\midrule
1x & 30  & Full Model & $\{0, 1, 2, 3, 4 \dots, 27, 28, 29\}$  \\
\midrule
\multirow{4}{*}{2x} & \multirow{4}{*}{15}  & Skip Top Layers & $\{0, 1, 2, 3, 4, 5, 6, 7, 8, 9, 10, 11, 12, 13, 14\}$   \\
~ & ~ & Skip Bottom Layers  &  $\{15, 16, 17, 18, 19, 20, 21, 22, 23, 24, 25, 26, 27, 28, 29\}$  \\
~ & ~ & \sysname &  $\{0, 2, 4, 6, 8, 10, 12, 14, 16, 18, 20, 22, 24, 26, 29\}$  \\
~ & ~ & Gready Searched Skipping & $\{0, 1, 2, 3, 4, 5, 12, 19, 21, 24, 25, 26, 27, 28, 29\}$ \\
\midrule
\multirow{4}{*}{3x} & \multirow{4}{*}{10}  & Skip Top Layers & $\{0, 1, 2, 3, 4, 5, 6, 7, 8, 9\}$   \\
~ & ~ & Skip Bottom Layers  &  $\{20, 21, 22, 23, 24, 25, 26, 27, 28, 29\}$  \\
~ & ~ & \sysname &  $\{0, 3, 6, 9, 12, 15, 18, 21, 24, 29\}$  \\
~ & ~ & Gready Searched Skipping & $\{0, 1, 2, 3, 19, 21, 25, 26, 27, 28, 29\}$ \\
\midrule
\multirow{4}{*}{5x} & \multirow{4}{*}{6}  & Skip Top Layers & $\{0, 1, 2, 3, 4, 5\}$   \\
~ & ~ & Skip Bottom Layers  &  $\{24, 25, 26, 27, 28, 29\}$  \\
~ & ~ & \sysname &  $\{0, 6, 12, 18, 24, 29\}$  \\
~ & ~ & Gready Searched Skipping & $\{0, 3, 19, 21, 28, 29\}$ \\

\bottomrule
\end{tabular}
}
\caption{The index of layers retained for different layer skipping strategies, using the BLOOMZ-7B as an example.
}
\label{tab:layer_to_keep}
\end{center}
\end{table*}
\subsection{Input-Independent Layer Skipping}
As summarized in Table~\ref{tab:comparision}, existing input-aware dynamic computation strategies still exhibit certain limitations in decoding efficiency. Although the more advanced Skip Decoding method can alleviate these issues to some extent, it still faces challenges with unstable speedup ratios and performance degradation due to unbalanced sub-network structures. Consequently, we propose a unified input-independent layer skipping strategy, named \sysname.

The fundamental reason why dynamic computation methods can bring about acceleration is by skipping the computation of some layers. How to determine the layer skipping choices of the model is a key difference among existing works. Although the current input-sensitive strategies are intuitive and have achieved certain effects, they also bring a series of computational efficiency issues. In this paper, we innovatively propose to decouple the model's layer skipping strategy from specific input samples, and the layer skipping choice is solely related to the target acceleration ratio. In this way, given the total number of model layers $N$ and the target acceleration ratio $r$, all samples execute the same layer skipping strategy and activate the same sub-layer network structure. This allows the model to easily support modern decoding acceleration strategies, such as batch decoding and KV caching. At the same time, the input-independent layer skipping strategy can ensure that the model will not be affected by the sample dynamic strategy during actual inference decoding, achieving stable, precise, and controllable acceleration effects.

\subsection{Which Layers to Skip?}
After we decide to execute the same layer skipping strategy for samples, how to specifically choose which layers to skip computation is also a key issue. As pointed out in existing works, not all layers in the model are equally important. Different layers of model parameters often undertake different responsibilities. For example, the lower layers of the model are responsible for input representation transformation, the middle layers are responsible for further abstracting the representation, and the top layers are often responsible for final prediction. Therefore, we believe that if the sub-network structure obtained after layer skipping can maintain a relatively balanced layer distribution, the negative impact of the layer skipping computation on the effect will be minimized. To verify this hypothesis, we adopted four different layer skipping strategies for experiments, namely, continuously skipping the top layers, continuously skipping the bottom layers, uniformly skipping the middle layers, and greedy search layer skipping strategy. The first strategy of skipping the top layers is adopted by the existing Early Exit series methods, the strategy of skipping the bottom layers is adopted by the more advanced Skip Decoding, the uniformly dispersed layer skipping is the strategy proposed in this paper, and the last greedy search strategy represents skipping the layer with the least impact on the effect each time, and repeating the above operation until the number of skipped layers meets the target acceleration ratio. Note that the greedy search strategy requires a large number of search experiments and is not practically efficient, so this strategy only represents an approximation to the ideal layer skipping situation and is only used as a strong baseline.

\begin{figure}[t!]
\begin{center}
     \scalebox{0.48}{           \includegraphics[width=1\textwidth]{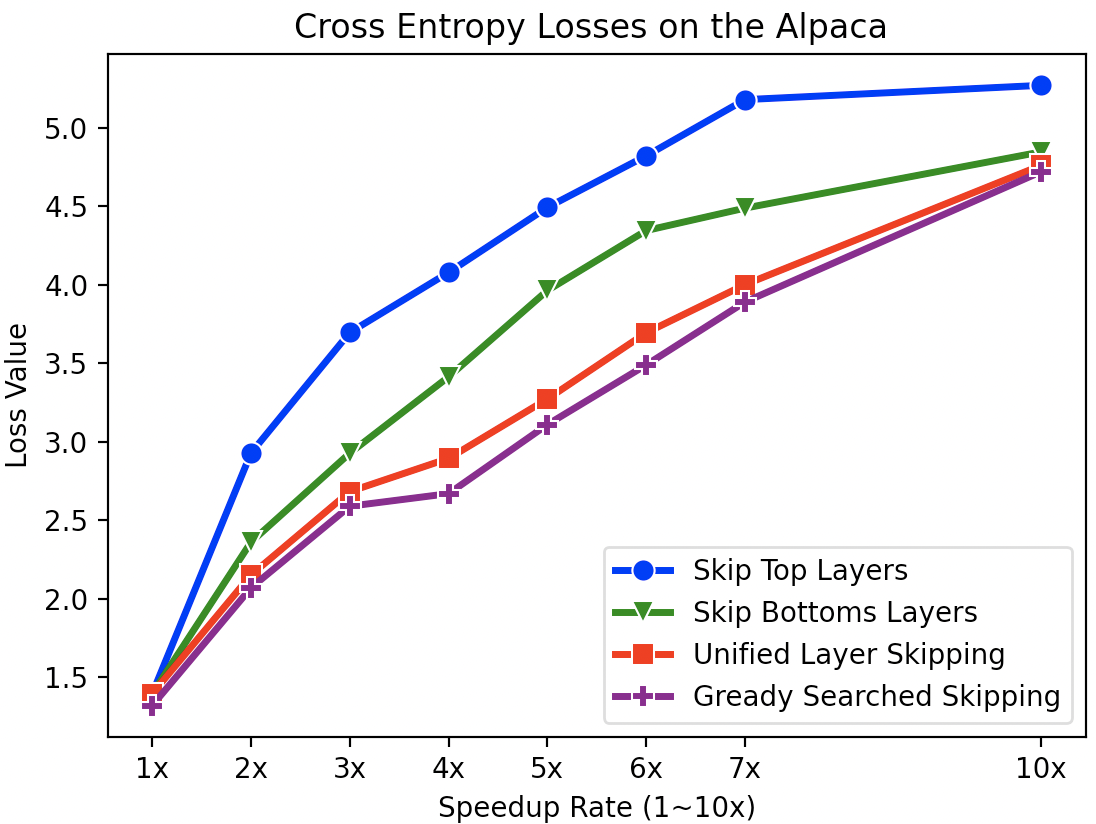}
      } 
      \caption{
       The values of cross-entropy loss corresponding to various layer-skipping strategies on the Alpaca dataset, where the smaller the value, the beter.
      } 
      \label{fig:plot_ce_loss}  
 \end{center}
\end{figure}

Here we use BLOOMZ-7B as a common example, perform supervised fine-tuning (SFT) experiments on the general task data Alpaca dataset, and list the cross-entropy loss values of different layer skipping strategies. The smaller the loss, the better the effect of the corresponding layer skipping strategy. In Table~\ref{tab:comparision}, we provide some examples of the number of layers selected by different layer skipping strategies under different target acceleration ratio requirements, and list the effects corresponding to different layer skipping strategies in Figure~\ref{fig:plot_ce_loss}. We can observe that the most commonly used strategy of skipping the top layers (the blue line in Figure~\ref{fig:plot_ce_loss}) has the largest loss of effect, because skipping the top prediction layer often leads to a significant decline in the model's prediction performance. Similarly, the strategy of skipping the bottom layers adopted by Skip Decoding (the green line in Figure~\ref{fig:plot_ce_loss}), although slightly better than the existing strategy of skipping the top layers, because the bottom representation of the model is responsible for some input feature transformation processing, continuously skipping the bottom layers will indirectly affect the final prediction effect. In contrast, the discrete layer skipping strategy proposed in this paper (the red line in Figure~\ref{fig:plot_ce_loss}) consistently improves performance across all acceleration ratio scenarios compared to the first two strategies, and achieves results comparable to the "theoretically optimal" greedy search strategy. This suggests that skipping the computation of the middle layers in a dispersed and uniform manner, resulting in a balanced sub-network structure, is not only more intuitive, but also demonstrates superior performance in empirical scenario.
\subsection{Formal Description}
Formally, the \sysname is illustrated in Algorithm 1. 
Specifically, the algorithm takes the total number of model layers $N$ and the target acceleration ratio $r$ as inputs. It outputs a set $S$ of retained layer ids after skipping layers. Initially, the set $S$ includes the bottom and top layer ids, {\em i.e.,} $0$ and $N-1$. 
The algorithm then calculates the number of layers to retain when conducting layer skipping, denoted as $M$, which is the floor division of 
$N$ by $r$. 
While the size of set $S$ is less than $M$, the algorithm iterates over the layers from $1$ to $N-2$. If the layer id 
$i$ is divisible by $r$, it will be added to the set $S$. This process continues until the size of set $S$ reaches $M$. Finally, the set $S$ is returned, representing the ids of the layers to be retained when conducting layer skipping.

Specifically, during the training phase, the model computes the full depth for the input prompt part, which is consistent with existing method~\cite{calm_2022,skipdecode_2023}. Subsequently, \sysname is only used in the response prediction phase during fine-tuning. For the inference phase, the model adopts the same settings as in the training phase.

\begin{algorithm}
\caption{\sysname}
\begin{algorithmic}[1]
\REQUIRE Total number of model layers $N$, target acceleration ratio $r$
\ENSURE Set of retained layer ids after skipping layers $S = \{\}$
\STATE \COMMENT{Include the bottom and top layer ids in the set $S$}
\STATE $S \gets \{0, N-1\}$
\STATE \COMMENT{Calculate the number of layers to retain after skipping layers}
\STATE $M \gets \lfloor \frac{N}{r} \rfloor$
\WHILE{$|S| < M$}
    \FOR{$i = 1$ to $N-2$}
        \IF{$i \mod r = 0$}
            \STATE \COMMENT{Add layer id $i$ to the set $S$ if it is divisible by $r$}
            \STATE $S \gets S \cup \{i\}$
        \ENDIF
    \ENDFOR
\ENDWHILE
\RETURN $S$
\end{algorithmic}
\end{algorithm}

\begin{table*}[t]
\begin{center}
\scalebox{0.8}{
\begin{tabular}{cclcccccc}
\toprule
\multirow{2}{*}{\textbf{\makecell{Target\\Speedup}}} &  \multirow{2}{*}{\textbf{\makecell{\#Target Activated \\ Layer}}}  & \multirow{2}{*}{\textbf{Approach}} &  \multirow{2}{*}{\textbf{\makecell{\#Actual Activated \\ Layer $\downarrow$}}} & \multicolumn{2}{c}{\textbf{Translation}} & \multicolumn{3}{c}{\textbf{Text Summarization}} \\
~ & ~ & ~ & ~ &\textbf{BELU$\uparrow$} & \textbf{COMET$\uparrow$} & \textbf{RG-1$\uparrow$} & \textbf{RG-2$\uparrow$} & \textbf{RG-L$\uparrow$}  \\
\midrule
1x & 30  & Full Model & 30.0  & 23.83 & 75.05 & 38.61 & 17.64  & 27.49  \\
\midrule
\multirow{4}{*}{2x} & \multirow{4}{*}{15}  & Early Exit & 18.2 &  17.82 & 65.27 & 29.83 &13.79 & 21.00 \\
~ &  ~ & SkipDecode & 21.5 & 20.70 & 69.77  & 34.41 &15.41 &24.09 \\
~ &  ~ & Unfied Skipping & 15.0 & \textbf{21.23}  & \textbf{70.81} & \textbf{34.67} & \textbf{16.13} & \textbf{25.24}  \\
\midrule

\multirow{4}{*}{3x}  & \multirow{4}{*}{10} & Early Exit& 12.1 &  12.35 & 56.31 & 21.03 &9.74 &16.06 \\
~ & ~ & SkipDecode & 13.7 &  15.52 & 59.94  & 25.90 & 12.21 &18.34 \\
~ & ~ & Unfied Skipping & 10.0 & \textbf{18.36} & \textbf{66.00}  & \textbf{30.71} &\textbf{14.38} &\textbf{21.54} \\
\midrule
\multirow{4}{*}{4x} & \multirow{4}{*}{7.5} &  Early Exit & 8.2  & 9.77 & 46.38  & 17.44 &8.34 &12.99 \\
~ & ~ & SkipDecode & 8.4 &  13.54 & 54.56  & 22.41 &10.47 &15.95 \\
~ & ~ & Unfied Skipping & 8.0  & \textbf{17.83}  & \textbf{65.54} & \textbf{27.86} & \textbf{13.62} &\textbf{21.12} \\
\midrule

\multirow{4}{*}{5x} & \multirow{4}{*}{6} & Early Exit& 7.1 &  10.02 & 46.95 &16.83 &9.38 &11.91 \\
~ & ~ & SkipDecode & 7.4 &  11.52 & 49.73 & 19.59 &9.38 &14.20  \\
~ & ~ & Unfied Skipping & 6.0  & \textbf{14.78} & \textbf{59.19} & \textbf{24.66} &\textbf{11.72} &\textbf{17.57} \\
\midrule

\multirow{4}{*}{10x} & \multirow{4}{*}{3} & Early Exit& 3.5 & 3.08 & 40.34 & 8.64 &3.07 &4.49   \\
~ & ~ & SkipDecode & 3.9  & 7.60 & 43.12 & 12.59 &5.84 &9.29   \\
~ & ~ & Unfied Skipping & 3.0 & \textbf{8.11} & \textbf{45.40} & \textbf{13.15} &\textbf{6.98} & \textbf{9.73}   \\

\bottomrule
\end{tabular}
}
\caption{Performance of various approaches on the BLOOMZ-7B backbone model at different target speedup ratios. The \textbf{bolded} scores correspond to the best performance under the same or comparable settings. "\#Target Activated Layer" means the average number of layers to be activated, and "\#Actual Activated layer" means the actual number of activated layers during the inference process of both tasks. "\sysnameshort" is short for "\sysname".
"RG-1", "RG-2" and "RG-1" are short for "ROUGE-1", "ROUGE-2" and "ROUGE-L", respectively.
$\downarrow$ indicates that smaller values are better, and $\uparrow$ indicates that larger values are better.
}
\vspace{-5pt}
\label{tab:bloomz}
\end{center}
\end{table*}

\section{Experiments and Results}
In this section, we commence with an overview of the training data, and then the specific training and inference settings.

\subsection{Datasets}
\label{sec:dataset}
we utilize the Alpaca dataset~\cite{alpaca} for both machine translation and text summarization.  For translation tasks, we use the WMT development sets from 2017 to 2020 as training data and fine-tune LLMs on data for Chinese-to-English, English-to-Chinese, German-to-English, and English-to-German translations following existing works~\cite{parrot_2023,post_ins_2023}. We evaluate translation performance using the test sets for these four directions in WMT-2022 and report the averaged BLEU and COMET scores. For text summarization, we use the CNN/DailyMail Dataset~\cite{prophetnet_2020,cnn1_2023}, a collection of over 300k unique English-language news articles. We conduct the evaluation on the standard test set with 11,490 samples and report the F1 scores of ROUGE-1, ROUGE-2, and ROUGE-L. Detaled setting are provied in Appendix~\ref{appendix:data}.

\subsection{Setups}
\paragraph{Training Setting.}
We take the polular BLOOMZ-7b~\cite{bloomz_2022} and LLaMa-13B~\cite{llama_2023} as our backbone models, each consists of 30 and 40 layers, respectively. We conduct instruction fine-tuning based on the wide-used Transformers framework~\cite{transformers_framework_2020}. On all experiments, we set the initial learning rate to 2e-5 and the rate of warmup to 0.03, and we tune the model with one epoch to mitigate the risk of overfitting. 
The maximum sequence length on the machine translation task and the text summarization task was set to 512 and 2,048, respectively.

\paragraph{Inference Settings.}
For all tasks, we set the batch size to 1 during inference to avoid the effect of padding side, unless specifically mentioned.
KV caching is used in all experiments and adapted to Early Exit with re-calculation following ~\citet{skipdecode_2023}.
As for the decoding strategies, we apply the beam search for all tasks, and set beam size to 4 for machine translation. While for the text summarization task, we have to decrease the beam size to 2, as encoding the long input sentences will consume a large portion of GPU memory.

\subsection{Comparison Systems}
\paragraph{Early Exit.} We select the most recent varient of Early Exit in our experiments, namely, the LITE~\cite{lite_2023}, which conduct instruction fine-tuning with an additional predicted losses from intermediate layers.
The speedup ratio of the LITE is regulated by adjusting the probability threshold for intermediate layer predictions.

\paragraph{SkipDecode.} Given a target speedup ratio, SkipDecode first specifies the upper and lower bounds of the model's executable layer, and then dynamically assigns different computational layers with the decoding position. 
We set up the above mentioned hyperparameters according to the original setup~\cite{skipdecode_2023}.

\section{Results and Analysis}

\subsection{Main Results}

The main results on BLOOMZ-7B are shown in Table~\ref{tab:bloomz}.
The Baseline Full Network serves as a reference point, showing the performance without any speedup techniques applied. 
The proposed \sysname consistently outperforms the strongly related Early Exit and Skip Decoding methods across different target acceleration ratios and task scenarios. For instance, at a 4x target speedup, it achieves an average improvement of 4.3 BLEU and 5.2 ROUGE-L compared to the strong baseline, Skip Decoding. 
In scenarios where the target acceleration demand is not very high, such as at 2x speedup, the proposed method can control the performance drop within a small range, around 2 BLEU or 2 ROUGE-L, compared to the non-accelerated baseline.
More importantly, given a target acceleration ratio and the number of layers to be used, \sysname allows for more precise control of the number of layers used in the actual generation process, thereby achieving a more stable acceleration effect.
On a larger-scale LLaMa-13B base model, similar experimental results are obtained in Table~\ref{tab:llama}, further corroborating the general applicability of the proposed method.

\subsection{Speedup in Actual Throughput}
Table \ref{tab:throughput} presents the average throughput (tokens/s) on one single NVIDIA A100 GPU (40G SXM) with the BLOOMZ-7B and the WMT-2022 test set, where the batch size is in the set {1,2,8}. The table compares the performance of different approaches, including Full-Network, Early Exit, SkipDecode, and our proposed \sysname, under different target speedup conditions (1x, 2x, and 5x) and batch sizes.
Experimental results indicate that, under different acceleration ratio scenarios, \sysname~can increase the throughput by 80\% to 290\% compared to the non-accelerated full-parameter model during online decoding ({\em i.e.,} when batch size is 1), without a significant drop in performance. Moreover, it further improves the throughput by approximately 30\% to 70\% compared to Early Exit and Skip Decoding methods, while ensuring the minimum performance loss at the same acceleration effect. During batch decoding, 
\sysname~still achieves similar conclusions when compared with Skip Decoding, namely, it coult increase the model throughput while ensuring the minimum performance loss.

\begin{table}[t]
\begin{center}
\scalebox{0.73}{
\begin{tabular}{ccllc}
\toprule
\multirow{2}{*}{\textbf{\makecell{Batch\\Size}}} & \multirow{2}{*}{\textbf{\makecell{Target\\Speedup}}} & \multirow{2}{*}{\textbf{Approach}} &  \multirow{2}{*}{\textbf{BLEU}}  & \multirow{2}{*}{\textbf{\makecell{Throughput\\(tokens/s)}}} \\
~ & ~ &  ~ & ~ & ~ \\
\midrule
\multirow{9}{*}{1} &  1x & Full-Network  & 23.81 & 51    \\
\cline{2-5}
~ & \multirow{4}{*}{2x}  & Early Exit & 17.82 & 73  \\ 
~ & ~ & SkipDecode & 20.70  &  67  \\
~ & ~ & \sysnameshort & \textbf{21.23}  & 90  \\
\cline{2-5}
~ & \multirow{4}{*}{5x}  & Early Exit & 10.02  & 167  \\
~ & ~ & SkipDecode & 11.52 & 154   \\
~ & ~ & \sysnameshort & \textbf{14.87}  & 201 \\
\midrule
\multirow{9}{*}{2} &  1x & Full-Network  & 23.81 & 92   \\
\cline{2-5}
~ & \multirow{4}{*}{2x}  & Early Exit & 17.82 & N/A  \\ 
~ & ~ & SkipDecode & 20.70  &  127  \\
~ & ~ & \sysnameshort & \textbf{21.23}  & 159  \\
\cline{2-5}
~ & \multirow{4}{*}{5x}  & Early Exit & 10.02 & N/A    \\
~ & ~ & SkipDecode & 11.52 &  285  \\
~ & ~ & \sysnameshort & \textbf{14.87} & 332   \\
\midrule
\multirow{9}{*}{8} &  1x & Full-Network  & 23.81 & 303   \\
\cline{2-5}
~ & \multirow{4}{*}{2x}  & Early Exit & 17.82 & N/A \\ 
~ & ~ & SkipDecode & 20.70  &  387  \\
~ & ~ & \sysnameshort & \textbf{21.23}  & 503 \\
\cline{2-5}
~ & \multirow{4}{*}{5x}  & Early Exit & 10.02 & N/A   \\
~ & ~ & SkipDecode & 11.52  & 864  \\
~ & ~ & \sysnameshort & \textbf{14.87} & 1004  \\
\bottomrule
\end{tabular}
}
\caption{Average throughput (tokens/s) on the BLOOMZ-7B on the WMT-2022 test set, where batch size $\in \{1, 2, 8\}$. 
`N/A' means this approach is not compatible with batch decoding. 
}
\vspace{-5pt}
\label{tab:throughput}
\end{center}
\end{table}

\subsection{In-depth Comparisons with LayerDrop}

Our method shares certain similarities with the layer pruning method, LayerDrop~\cite{layerdrop_2019}, and in this section, we will conduct an in-depth comparison between these two method. LayerDrop consists of two parts: firstly, it randomly drops some layers during the training phase, similar to dropout~\cite{dropout_2014}, thereby making the model be resilient to dropping some layers during decoding. 
Subsequently, it selects some layers for computation based on the target acceleration ratio.
The fundamental difference between LayerDrop and \sysname~lies in the fact that LayerDrop, as a structured pruning method, cannot distinguish between the input prompt and response prediction parts. This leads to the pruned sub-model being unable to perform full-depth feed-forward computation on the input prompt part, resulting in a significant decline in the performance of LayerDrop when directly applied to decoder-only structured LLMs. We improve LayerDrop and adapt it to the decoder-only structure with full-depth computation on the input prompt part and pruning only for the response prediction part. The results in Table~\ref{tab:layer_to_keep} show that this improved version of LayerDrop can alleviate its performance decline issue. More importantly, after combining LayerDrop with the \sysname~proposed in this paper, we observe further performance gains, demonstrating the complementarity and combinability of the two methods.

\begin{table}[t]
\begin{center}
\scalebox{0.8}{
\begin{tabular}{clc}
\toprule
\multirow{2}{*}{\textbf{\makecell{Target\\Speedup}}}  & \multirow{2}{*}{\textbf{Approach}}  &  \multirow{2}{*}{\textbf{BLEU}} \\ 
~ & ~ & ~  \\
\midrule
1x  & Full Model & 23.83  \\
\midrule
\multirow{3}{*}{2x} &  LayerDrop & 17.12 \\
~ &   LayerDrop + Deep Encoding & 19.28 \\
~ & \sysnameshort & 21.23 \\
~ &   LayerDrop + \sysnameshort & \textbf{22.19} \\
\midrule
\multirow{3}{*}{5x} &  LayerDrop & 8.76 \\
~ &   LayerDrop + Deep Encoding & 12.45 \\
~ & \sysnameshort & 14.78 \\
~ &   LayerDrop + \sysnameshort & \textbf{15.09} \\
\bottomrule
\end{tabular}
}
\caption{Machine translation performance (averaged BLEU) of serveral approaches on the BLOOMZ-7B backbone model on the WMT-2022 test set. 
}
\vspace{-5pt}
\label{tab:layerdrop}
\end{center}
\end{table}

\section{Conclusion}
In conclusion, our proposed \sysname~strategy addresses the limitations of existing dynamic computation methods for LLMs, which cannot guarantee a stable and precise acceleration effect due to the assignment of different computational budgets to different samples during decoding. Furthermore, our strategy mitigates the performance degeneration caused by the drastic change in the model’s layer-wise representations. The \sysname~strategy significantly enhances both the inference performance and the actual model throughput. Moreover, its independence from input samples allows it to naturally support popular acceleration techniques such as batch decoding and KV caching, demonstrating its practicality for real-world applications.

\section*{Limitation}
Despite the promising results, our study has some limitations. The \sysname~strategy, while effective, is based on a fixed target speedup ratio, which may not be optimal for all scenarios. The strategy's performance may vary depending on the complexity and diversity of the tasks at hand. Lastly, while we have demonstrated the effectiveness of our approach in machine translation and text summarization tasks, further research is needed to validate its applicability and performance in other NLP tasks and real-world applications.

\bibliography{anthology,custom}

\begin{thebibliography}{52}
\expandafter\ifx\csname natexlab\endcsname\relax\def\natexlab#1{#1}\fi

\bibitem[{Aher et~al.(2023)Aher, Arriaga, and Kalai}]{follow_instructgpt6}
Gati~V Aher, Rosa~I Arriaga, and Adam~Tauman Kalai. 2023.
\newblock Using large language models to simulate multiple humans and replicate human subject studies.
\newblock In \emph{International Conference on Machine Learning}, pages 337--371. PMLR.

\bibitem[{Ahn et~al.(2022)Ahn, Brohan, Brown, Chebotar, Cortes, David, Finn, Fu, Gopalakrishnan, Hausman et~al.}]{follow_instructgpt4}
Michael Ahn, Anthony Brohan, Noah Brown, Yevgen Chebotar, Omar Cortes, Byron David, Chelsea Finn, Chuyuan Fu, Keerthana Gopalakrishnan, Karol Hausman, et~al. 2022.
\newblock Do as i can, not as i say: Grounding language in robotic affordances.
\newblock \emph{arXiv preprint arXiv:2204.01691}.

\bibitem[{Ayd{\i}n and Karaarslan(2023)}]{chatgpt6}
{\"O}mer Ayd{\i}n and Enis Karaarslan. 2023.
\newblock Is chatgpt leading generative ai? what is beyond expectations?
\newblock \emph{What is beyond expectations}.

\bibitem[{Brooks et~al.(2023)Brooks, Holynski, and Efros}]{follow_instructgpt1}
Tim Brooks, Aleksander Holynski, and Alexei~A Efros. 2023.
\newblock Instructpix2pix: Learning to follow image editing instructions.
\newblock In \emph{Proceedings of the IEEE/CVF Conference on Computer Vision and Pattern Recognition}, pages 18392--18402.

\bibitem[{Chung et~al.(2022)Chung, Hou, Longpre, Zoph, Tay, Fedus, Li, Wang, Dehghani, Brahma et~al.}]{follow_instructgpt2}
Hyung~Won Chung, Le~Hou, Shayne Longpre, Barret Zoph, Yi~Tay, William Fedus, Eric Li, Xuezhi Wang, Mostafa Dehghani, Siddhartha Brahma, et~al. 2022.
\newblock Scaling instruction-finetuned language models.
\newblock \emph{arXiv preprint arXiv:2210.11416}.

\bibitem[{Dehghani et~al.(2019)Dehghani, Gouws, Vinyals, Uszkoreit, and Łukasz Kaiser}]{UT_2019}
Mostafa Dehghani, Stephan Gouws, Oriol Vinyals, Jakob Uszkoreit, and Łukasz Kaiser. 2019.
\newblock Universal transformers.
\newblock In \emph{ICLR}.

\bibitem[{Del~Corro et~al.(2023)Del~Corro, Del~Giorno, Agarwal, Yu, Awadallah, and Mukherjee}]{skipdecode_2023}
Luciano Del~Corro, Allie Del~Giorno, Sahaj Agarwal, Bin Yu, Ahmed Awadallah, and Subhabrata Mukherjee. 2023.
\newblock Skipdecode: Autoregressive skip decoding with batching and caching for efficient llm inference.
\newblock \emph{arXiv preprint arXiv:2307.02628}.

\bibitem[{Dettmers et~al.(2022)Dettmers, Lewis, Belkada, and Zettlemoyer}]{llm_quant_2022}
Tim Dettmers, Mike Lewis, Younes Belkada, and Luke Zettlemoyer. 2022.
\newblock Llm. int8 (): 8-bit matrix multiplication for transformers at scale.
\newblock \emph{arXiv preprint arXiv:2208.07339}.

\bibitem[{Elbayad et~al.(2019)Elbayad, Gu, Grave, and Auli}]{da_transformer_2019}
Maha Elbayad, Jiatao Gu, Edouard Grave, and Michael Auli. 2019.
\newblock Depth-adaptive transformer.
\newblock \emph{arXiv preprint arXiv:1910.10073}.

\bibitem[{Fan et~al.(2020)Fan, Grave, and Joulin}]{layerdrop_2019}
Angela Fan, Edouard Grave, and Armand Joulin. 2020.
\newblock Reducing transformer depth on demand with structured dropout.
\newblock In \emph{{ICLR} 2020}.

\bibitem[{Figurnov et~al.(2017{\natexlab{a}})Figurnov, Collins, Zhu, Zhang, Huang, Vetrov, and Salakhutdinov}]{ACT_resnet_2017}
Michael Figurnov, Maxwell~D. Collins, Yukun Zhu, Li~Zhang, Jonathan Huang, Dmitry Vetrov, and Ruslan Salakhutdinov. 2017{\natexlab{a}}.
\newblock Spatially adaptive computation time for residual networks.
\newblock In \emph{CVPR}.

\bibitem[{Figurnov et~al.(2017{\natexlab{b}})Figurnov, Sobolev, and Vetrov}]{ee_2017}
Michael Figurnov, Artem Sobolev, and Dmitry Vetrov. 2017{\natexlab{b}}.
\newblock Probabilistic adaptive computation time.
\newblock \emph{arXiv preprint arXiv:1712.00386}.

\bibitem[{Graves(2016)}]{ACT_2016}
Alex Graves. 2016.
\newblock Adaptive computation time for recurrent neural networks.
\newblock \emph{arXiv}.

\bibitem[{Hill-Yardin et~al.(2023)Hill-Yardin, Hutchinson, Laycock, and Spencer}]{chatgpt4}
Elisa~L Hill-Yardin, Mark~R Hutchinson, Robin Laycock, and Sarah~J Spencer. 2023.
\newblock A chat (gpt) about the future of scientific publishing.
\newblock \emph{Brain Behav Immun}, 110:152--154.

\bibitem[{Jiao et~al.(2023{\natexlab{a}})Jiao, tse Huang, Wang, Wang, Shi, and Tu}]{parrot_2023}
Wenxiang Jiao, Jen tse Huang, Wenxuan Wang, Xing Wang, Shuming Shi, and Zhaopeng Tu. 2023{\natexlab{a}}.
\newblock Parrot: Translating during chat using large language models.
\newblock In \emph{ArXiv}.

\bibitem[{Jiao et~al.(2023{\natexlab{b}})Jiao, tse Huang, Wang, Wang, Shi, and Tu}]{jiao2023parrot}
Wenxiang Jiao, Jen tse Huang, Wenxuan Wang, Xing Wang, Shuming Shi, and Zhaopeng Tu. 2023{\natexlab{b}}.
\newblock Parrot: Translating during chat using large language models.
\newblock In \emph{ArXiv}.

\bibitem[{Jiao et~al.(2023{\natexlab{c}})Jiao, Wang, Huang, Wang, and Tu}]{chatgpt2}
Wenxiang Jiao, Wenxuan Wang, Jen-tse Huang, Xing Wang, and Zhaopeng Tu. 2023{\natexlab{c}}.
\newblock Is chatgpt a good translator? a preliminary study.
\newblock \emph{arXiv preprint arXiv:2301.08745}.

\bibitem[{Kasneci et~al.(2023)Kasneci, Se{\ss}ler, K{\"u}chemann, Bannert, Dementieva, Fischer, Gasser, Groh, G{\"u}nnemann, H{\"u}llermeier et~al.}]{chatgpt3}
Enkelejda Kasneci, Kathrin Se{\ss}ler, Stefan K{\"u}chemann, Maria Bannert, Daryna Dementieva, Frank Fischer, Urs Gasser, Georg Groh, Stephan G{\"u}nnemann, Eyke H{\"u}llermeier, et~al. 2023.
\newblock Chatgpt for good? on opportunities and challenges of large language models for education.
\newblock \emph{Learning and Individual Differences}, 103:102274.

\bibitem[{Li et~al.(2023)Li, Zhang, and Chen}]{llm_kd_2023}
Lei Li, Yongfeng Zhang, and Li~Chen. 2023.
\newblock Prompt distillation for efficient llm-based recommendation.
\newblock In \emph{Proceedings of the 32nd ACM International Conference on Information and Knowledge Management}, pages 1348--1357.

\bibitem[{Lin et~al.(2023{\natexlab{a}})Lin, Tang, Tang, Yang, Dang, and Han}]{llm_quant1_2023}
Ji~Lin, Jiaming Tang, Haotian Tang, Shang Yang, Xingyu Dang, and Song Han. 2023{\natexlab{a}}.
\newblock Awq: Activation-aware weight quantization for llm compression and acceleration.
\newblock \emph{arXiv preprint arXiv:2306.00978}.

\bibitem[{Lin et~al.(2023{\natexlab{b}})Lin, Gong, Shen, Wu, Fan, Lin, Duan, and Chen}]{cnn3_2023}
Zhenghao Lin, Yeyun Gong, Yelong Shen, Tong Wu, Zhihao Fan, Chen Lin, Nan Duan, and Weizhu Chen. 2023{\natexlab{b}}.
\newblock Text generation with diffusion language models: A pre-training approach with continuous paragraph denoise.
\newblock In \emph{International Conference on Machine Learning}, pages 21051--21064. PMLR.

\bibitem[{Liu et~al.(2021{\natexlab{a}})Liu, Meng, Chen, Xu, and Zhou}]{ss_decoding_2021}
Yijin Liu, Fandong Meng, Yufeng Chen, Jinan Xu, and Jie Zhou. 2021{\natexlab{a}}.
\newblock Scheduled sampling based on decoding steps for neural machine translation.
\newblock In \emph{Proceedings of the 2021 Conference on Empirical Methods in Natural Language Processing}, pages 3285--3296.

\bibitem[{Liu et~al.(2021{\natexlab{b}})Liu, Meng, Zhou, Chen, and Xu}]{faster_da_2021}
Yijin Liu, Fandong Meng, Jie Zhou, Yufeng Chen, and Jinan Xu. 2021{\natexlab{b}}.
\newblock Faster depth-adaptive transformers.
\newblock In \emph{Proceedings of the AAAI Conference on Artificial Intelligence}, volume~35, pages 13424--13432.

\bibitem[{Liu et~al.(2023{\natexlab{a}})Liu, Zeng, Meng, and Zhou}]{post_ins_2023}
Yijin Liu, Xianfeng Zeng, Fandong Meng, and Jie Zhou. 2023{\natexlab{a}}.
\newblock Instruction position matters in sequence generation with large language models.
\newblock \emph{arXiv preprint arXiv:2308.12097}.

\bibitem[{Liu et~al.(2023{\natexlab{b}})Liu, Oguz, Zhao, Chang, Stock, Mehdad, Shi, Krishnamoorthi, and Chandra}]{llm_quant2_2023}
Zechun Liu, Barlas Oguz, Changsheng Zhao, Ernie Chang, Pierre Stock, Yashar Mehdad, Yangyang Shi, Raghuraman Krishnamoorthi, and Vikas Chandra. 2023{\natexlab{b}}.
\newblock Llm-qat: Data-free quantization aware training for large language models.
\newblock \emph{arXiv preprint arXiv:2305.17888}.

\bibitem[{Liu et~al.(2021{\natexlab{c}})Liu, Xu, Wang, Darrell, and Shelhamer}]{ee_2021}
Zhuang Liu, Zhiqiu Xu, Hung-Ju Wang, Trevor Darrell, and Evan Shelhamer. 2021{\natexlab{c}}.
\newblock Anytime dense prediction with confidence adaptivity.
\newblock \emph{arXiv preprint arXiv:2104.00749}.

\bibitem[{Ma et~al.(2023)Ma, Fang, and Wang}]{llm_prune_2023}
Xinyin Ma, Gongfan Fang, and Xinchao Wang. 2023.
\newblock Llm-pruner: On the structural pruning of large language models.
\newblock \emph{arXiv preprint arXiv:2305.11627}.

\bibitem[{Muennighoff et~al.(2022)Muennighoff, Wang, Sutawika, Roberts, Biderman, Scao, Bari, Shen, Yong, Schoelkopf et~al.}]{bloomz_2022}
Niklas Muennighoff, Thomas Wang, Lintang Sutawika, Adam Roberts, Stella Biderman, Teven~Le Scao, M~Saiful Bari, Sheng Shen, Zheng-Xin Yong, Hailey Schoelkopf, et~al. 2022.
\newblock Crosslingual generalization through multitask finetuning.
\newblock \emph{arXiv preprint arXiv:2211.01786}.

\bibitem[{Muhamed et~al.(2021)Muhamed, Keivanloo, Perera, Mracek, Xu, Cui, Rajagopalan, Zeng, and Chilimbi}]{llm_kd_2021}
Aashiq Muhamed, Iman Keivanloo, Sujan Perera, James Mracek, Yi~Xu, Qingjun Cui, Santosh Rajagopalan, Belinda Zeng, and Trishul Chilimbi. 2021.
\newblock Ctr-bert: Cost-effective knowledge distillation for billion-parameter teacher models.
\newblock In \emph{NeurIPS Efficient Natural Language and Speech Processing Workshop}.

\bibitem[{Ouyang et~al.(2022)Ouyang, Wu, Jiang, Almeida, Wainwright, Mishkin, Zhang, Agarwal, Slama, Ray et~al.}]{instructgpt_2022}
Long Ouyang, Jeffrey Wu, Xu~Jiang, Diogo Almeida, Carroll Wainwright, Pamela Mishkin, Chong Zhang, Sandhini Agarwal, Katarina Slama, Alex Ray, et~al. 2022.
\newblock Training language models to follow instructions with human feedback.
\newblock \emph{Advances in Neural Information Processing Systems}, 35:27730--27744.

\bibitem[{Qi et~al.(2020)Qi, Yan, Gong, Liu, Duan, Chen, Zhang, and Zhou}]{prophetnet_2020}
Weizhen Qi, Yu~Yan, Yeyun Gong, Dayiheng Liu, Nan Duan, Jiusheng Chen, Ruofei Zhang, and Ming Zhou. 2020.
\newblock \href {https://www.aclweb.org/anthology/2020.findings-emnlp.217} {{P}rophet{N}et: Predicting future n-gram for sequence-to-{S}equence{P}re-training}.
\newblock In \emph{Findings of the Association for Computational Linguistics: EMNLP 2020}, pages 2401--2410, Online. Association for Computational Linguistics.

\bibitem[{Schuster et~al.(2022)Schuster, Fisch, Gupta, Dehghani, Bahri, Tran, Tay, and Metzler}]{calm_2022}
Tal Schuster, Adam Fisch, Jai Gupta, Mostafa Dehghani, Dara Bahri, Vinh Tran, Yi~Tay, and Donald Metzler. 2022.
\newblock \href {https://proceedings.neurips.cc/paper_files/paper/2022/file/6fac9e316a4ae75ea244ddcef1982c71-Paper-Conference.pdf} {Confident adaptive language modeling}.
\newblock In \emph{Advances in Neural Information Processing Systems}, volume~35, pages 17456--17472. Curran Associates, Inc.

\bibitem[{See et~al.(2017)See, Liu, and Manning}]{cnndw_dataset_2017}
Abigail See, Peter~J. Liu, and Christopher~D. Manning. 2017.
\newblock \href {https://doi.org/10.18653/v1/P17-1099} {Get to the point: Summarization with pointer-generator networks}.
\newblock In \emph{Proceedings of the 55th Annual Meeting of the Association for Computational Linguistics (Volume 1: Long Papers)}, pages 1073--1083, Vancouver, Canada. Association for Computational Linguistics.

\bibitem[{{\v{S}}lapeta(2023)}]{chatgpt5}
Jan {\v{S}}lapeta. 2023.
\newblock Are chatgpt and other pretrained language models good parasitologists?
\newblock \emph{Trends in Parasitology}.

\bibitem[{Srivastava et~al.(2014)Srivastava, Hinton, Krizhevsky, Sutskever, and Salakhutdinov}]{dropout_2014}
Nitish Srivastava, Geoffrey Hinton, Alex Krizhevsky, Ilya Sutskever, and Ruslan Salakhutdinov. 2014.
\newblock Dropout: a simple way to prevent neural networks from overfitting.
\newblock \emph{The journal of machine learning research}, 15(1):1929--1958.

\bibitem[{Sun et~al.(2023)Sun, Liu, Bair, and Kolter}]{llm_prune1_2023}
Mingjie Sun, Zhuang Liu, Anna Bair, and J~Zico Kolter. 2023.
\newblock A simple and effective pruning approach for large language models.
\newblock \emph{arXiv preprint arXiv:2306.11695}.

\bibitem[{Tang et~al.(2023)Tang, Wang, Wang, Chen, Gao, and Qian}]{cnn1_2023}
Moming Tang, Chengyu Wang, Jianing Wang, Cen Chen, Ming Gao, and Weining Qian. 2023.
\newblock Parasum: Contrastive paraphrasing for low-resource extractive text summarization.
\newblock In \emph{International Conference on Knowledge Science, Engineering and Management}, pages 106--119. Springer.

\bibitem[{Taori et~al.(2023)Taori, Gulrajani, Zhang, Dubois, Li, Guestrin, Liang, and Hashimoto}]{alpaca}
Rohan Taori, Ishaan Gulrajani, Tianyi Zhang, Yann Dubois, Xuechen Li, Carlos Guestrin, Percy Liang, and Tatsunori~B. Hashimoto. 2023.
\newblock Stanford alpaca: An instruction-following llama model.
\newblock \url{https://github.com/tatsu-lab/stanford_alpaca}.

\bibitem[{Teerapittayanon et~al.(2016)Teerapittayanon, McDanel, and Kung}]{ee_2016}
Surat Teerapittayanon, Bradley McDanel, and Hsiang-Tsung Kung. 2016.
\newblock Branchynet: Fast inference via early exiting from deep neural networks.
\newblock In \emph{2016 23rd international conference on pattern recognition (ICPR)}, pages 2464--2469. IEEE.

\bibitem[{Touvron et~al.(2023)Touvron, Lavril, Izacard, Martinet, Lachaux, Lacroix, Rozi{\`e}re, Goyal, Hambro, Azhar et~al.}]{llama_2023}
Hugo Touvron, Thibaut Lavril, Gautier Izacard, Xavier Martinet, Marie-Anne Lachaux, Timoth{\'e}e Lacroix, Baptiste Rozi{\`e}re, Naman Goyal, Eric Hambro, Faisal Azhar, et~al. 2023.
\newblock Llama: Open and efficient foundation language models.
\newblock \emph{arXiv preprint arXiv:2302.13971}.

\bibitem[{Varshney et~al.(2023)Varshney, Chatterjee, Parmar, and Baral}]{lite_2023}
Neeraj Varshney, Agneet Chatterjee, Mihir Parmar, and Chitta Baral. 2023.
\newblock Accelerating llama inference by enabling intermediate layer decoding via instruction tuning with lite.
\newblock \emph{arXiv e-prints}, pages arXiv--2310.

\bibitem[{Wang et~al.(2023)Wang, Liang, Meng, Shi, Li, Xu, Qu, and Zhou}]{chatgpt1}
Jiaan Wang, Yunlong Liang, Fandong Meng, Haoxiang Shi, Zhixu Li, Jinan Xu, Jianfeng Qu, and Jie Zhou. 2023.
\newblock Is chatgpt a good nlg evaluator? a preliminary study.
\newblock \emph{arXiv preprint arXiv:2303.04048}.

\bibitem[{Wang et~al.(2022)Wang, Kordi, Mishra, Liu, Smith, Khashabi, and Hajishirzi}]{self_instruct_2022}
Yizhong Wang, Yeganeh Kordi, Swaroop Mishra, Alisa Liu, Noah~A Smith, Daniel Khashabi, and Hannaneh Hajishirzi. 2022.
\newblock Self-instruct: Aligning language model with self generated instructions.
\newblock \emph{arXiv preprint arXiv:2212.10560}.

\bibitem[{Wei et~al.(2021)Wei, Bosma, Zhao, Guu, Yu, Lester, Du, Dai, and Le}]{follow_instructgpt5}
Jason Wei, Maarten Bosma, Vincent~Y Zhao, Kelvin Guu, Adams~Wei Yu, Brian Lester, Nan Du, Andrew~M Dai, and Quoc~V Le. 2021.
\newblock Finetuned language models are zero-shot learners.
\newblock \emph{arXiv preprint arXiv:2109.01652}.

\bibitem[{Wei et~al.(2022)Wei, Tay, Bommasani, Raffel, Zoph, Borgeaud, Yogatama, Bosma, Zhou, Metzler et~al.}]{follow_instructgpt3}
Jason Wei, Yi~Tay, Rishi Bommasani, Colin Raffel, Barret Zoph, Sebastian Borgeaud, Dani Yogatama, Maarten Bosma, Denny Zhou, Donald Metzler, et~al. 2022.
\newblock Emergent abilities of large language models.
\newblock \emph{arXiv preprint arXiv:2206.07682}.

\bibitem[{Wolf et~al.(2020)Wolf, Debut, Sanh, Chaumond, Delangue, Moi, Cistac, Rault, Louf, Funtowicz, Davison, Shleifer, von Platen, Ma, Jernite, Plu, Xu, Scao, Gugger, Drame, Lhoest, and Rush}]{transformers_framework_2020}
Thomas Wolf, Lysandre Debut, Victor Sanh, Julien Chaumond, Clement Delangue, Anthony Moi, Pierric Cistac, Tim Rault, Rémi Louf, Morgan Funtowicz, Joe Davison, Sam Shleifer, Patrick von Platen, Clara Ma, Yacine Jernite, Julien Plu, Canwen Xu, Teven~Le Scao, Sylvain Gugger, Mariama Drame, Quentin Lhoest, and Alexander~M. Rush. 2020.
\newblock \href {https://www.aclweb.org/anthology/2020.emnlp-demos.6} {Transformers: State-of-the-art natural language processing}.
\newblock In \emph{Proceedings of the 2020 Conference on Empirical Methods in Natural Language Processing: System Demonstrations}, pages 38--45, Online. Association for Computational Linguistics.

\bibitem[{Xiao et~al.(2023)Xiao, Lin, Seznec, Wu, Demouth, and Han}]{llm_quant_2023}
Guangxuan Xiao, Ji~Lin, Mickael Seznec, Hao Wu, Julien Demouth, and Song Han. 2023.
\newblock Smoothquant: Accurate and efficient post-training quantization for large language models.
\newblock In \emph{International Conference on Machine Learning}, pages 38087--38099. PMLR.

\bibitem[{Zafrir et~al.(2019)Zafrir, Boudoukh, Izsak, and Wasserblat}]{quant_2019}
Ofir Zafrir, Guy Boudoukh, Peter Izsak, and Moshe Wasserblat. 2019.
\newblock Q8bert: Quantized 8bit bert.
\newblock In \emph{2019 Fifth Workshop on Energy Efficient Machine Learning and Cognitive Computing-NeurIPS Edition (EMC2-NIPS)}, pages 36--39. IEEE.

\bibitem[{Zeng et~al.(2023)Zeng, Meng, Yin, and Zhou}]{zeng2023tim}
Jiali Zeng, Fandong Meng, Yongjing Yin, and Jie Zhou. 2023.
\newblock \href {https://arxiv.org/pdf/2307.04408.pdf} {Tim: Teaching lm to translate with comparison}.
\newblock In \emph{ArXiv}.

\bibitem[{Zhang et~al.(2023{\natexlab{a}})Zhang, Liu, and Zhang}]{cnn2_2023}
Haopeng Zhang, Xiao Liu, and Jiawei Zhang. 2023{\natexlab{a}}.
\newblock Diffusum: Generation enhanced extractive summarization with diffusion.
\newblock \emph{arXiv preprint arXiv:2305.01735}.

\bibitem[{Zhang et~al.(2020)Zhang, Zhou, Zhao, and Zong}]{accum_error_2020}
Jiajun Zhang, Long Zhou, Yang Zhao, and Chengqing Zong. 2020.
\newblock Synchronous bidirectional inference for neural sequence generation.
\newblock \emph{Artificial Intelligence}, 281:103234.

\bibitem[{Zhang et~al.(2023{\natexlab{b}})Zhang, Xiao, Liu, Dou, and Nie}]{llm_kd1_2023}
Peitian Zhang, Shitao Xiao, Zheng Liu, Zhicheng Dou, and Jian-Yun Nie. 2023{\natexlab{b}}.
\newblock Retrieve anything to augment large language models.
\newblock \emph{arXiv preprint arXiv:2310.07554}.

\end{thebibliography}

\appendix

\section{Detailed Dataset Settings}
\label{appendix:data}

\paragraph{Alpaca.} 
The Alpaca dataset, released by Stanford~\cite{alpaca}, is widely used for instruction following tasks.
It is constructed by the self-instruct data generation pipline~\cite{self_instruct_2022}, using the text-davinci-003 model to generate high-quality instruction-following data.
Following~\citet{post_ins_2023}, we relocate the positions of inputs and task instructions to emphasizes the instructions during generation in all experiments.
Please note that the Alpaca dataset is a general instruction-following task dataset, and we utilize it for both the machine translation and text summarization tasks.

\paragraph{WMT Datasets.} 
We use the WMT development sets from 2017 to 2020 as high-quality translation training data, following existing settings~\cite{jiao2023parrot,zeng2023tim,post_ins_2023}.

To facilitate comparison, we follow the settings of existing methods~\cite{jiao2023parrot,zeng2023tim} and fine-tune LLMs on data for three languages and four translation directions: Chinese-to-English, English-to-Chinese, German-to-English, and English-to-German. There are totally about 51k of sentence pairs for instruction fine-tuning.
The test sets for these four directions in WMT-2022 are used to evaluate translation performance. 
We report the averaged BLEU scores calculated with SacreBLEU\footnote{\url{ https://github.com/mjpost/sacrebleu}} and averaged COMET scores on the recent \textit{wmt22-cometkiwi-da}\footnote{\url{https://github.com/Unbabel/COMET}} for the above four translation directions.

\paragraph{CNN/DailyMail.}
The popular CNN/DailyMail Dataset~\cite{cnndw_dataset_2017} is a collection of English-language news articles, comprising slightly over 300k unique articles authored by journalists from CNN and the Daily Mail.
The average sentence length of the source text of these data is about 665 words, or about a thousand tokens, which served as a widely used benchmark for long text summarization~\cite{cnn1_2023,cnn2_2023,cnn3_2023}.
We follow the pre-processing and post-processing scripts of existing studies~\cite{prophetnet_2020}.
We use the CNN/DailyMail dataset only for the text summarization task and conduct the evaluation on the standard test set with 11,490 samples.
We report the F1 scores of ROUGE-1, ROUGE-2, and ROUGE-L following existing studies~\cite{cnn1_2023,prophetnet_2020}.

\section{Performance on the LLaMA-13B backbone}

\begin{table*}[t]
\begin{center}
\scalebox{0.8}{
\begin{tabular}{cclcccccc}
\toprule
\multirow{2}{*}{\textbf{\makecell{Target\\Speedup}}} &  \multirow{2}{*}{\textbf{\makecell{\#Target Activated \\ Layer}}}  & \multirow{2}{*}{\textbf{Approach}} &  \multirow{2}{*}{\textbf{\makecell{\#Actual Activated \\ Layer $\downarrow$}}} & \multicolumn{2}{c}{\textbf{Translation}} & \multicolumn{3}{c}{\textbf{Text Summarization}} \\
~ & ~ & ~ & ~ &\textbf{BELU$\uparrow$} & \textbf{COMET$\uparrow$} & \textbf{RG-1$\uparrow$} & \textbf{RG-2$\uparrow$} & \textbf{RG-L$\uparrow$}  \\
\midrule
1x & 40  & Full Model & 40.0  & 27.84 & 80.83 & 40.13 & 19.27  & 29.24  \\
\midrule
\multirow{4}{*}{2x} & \multirow{4}{*}{20}  & Early Exit & 24.1 &  25.41 & 76.27 & 38.05 & 18.78 & 27.43 \\
~ &  ~ & SkipDecode & 27.8 & 25.68 & 77.10  & 37.99 & 19.24 & 27.35 \\
~ &  ~ & Unfied Skipping & 20.0 & \textbf{26.93}  & \textbf{78.74} & \textbf{39.03} & \textbf{18.81} & \textbf{28.65}  \\
\midrule

\multirow{4}{*}{3x}  & \multirow{4}{*}{13} & Early Exit& 15.6 & 22.71  & 73.22 & 34.04 & 16.46 & 24.02 \\
~ & ~ & SkipDecode & 17.3 &  23.49 & 74.18  & 35.60 & 17.36 & \textbf{26.41} \\
~ & ~ & Unfied Skipping & 13.0 & \textbf{24.82} & \textbf{75.43} & \textbf{37.52} & \textbf{18.50} & 26.37  \\
\midrule
\multirow{4}{*}{4x} & \multirow{4}{*}{10} &  Early Exit & 10.6  & 19.85  & 72.19  & 29.38 & 13.96 & 22.36\\
~ & ~ & SkipDecode & 11.1 &  21.73 & 73.03  & 31.44 & 15.54 & 23.49\\
~ & ~ & Unfied Skipping & 10.0  & \textbf{24.24} & \textbf{75.05} & \textbf{35.37} & \textbf{17.61} & \textbf{26.21} \\
\midrule

\multirow{4}{*}{5x} & \multirow{4}{*}{8} & Early Exit& 9.1 &  17.17 & 70.92 &  25.71 & 12.88 & 19.18  \\
~ & ~ & SkipDecode & 9.3 &  16.53 & 71.94 & 24.95 & 13.16 & 18.00 \\
~ & ~ & Unfied Skipping & 8.0  & \textbf{21.55} & \textbf{72.64} & \textbf{32.40} & \textbf{16.33} & \textbf{23.29} \\
\midrule

\multirow{4}{*}{10x} & \multirow{4}{*}{4} & Early Exit& 4.3 & 9.72 & 61.03  & 14.53 & 7.13 & 11.75 \\
~ & ~ & SkipDecode & 4.8  & 12.15 & 65.31 & 17.76 & 9.64 & 13.62 \\
~ & ~ & Unfied Skipping & 4.0 & \textbf{15.67} & \textbf{68.27} & \textbf{23.52} & \textbf{12.48} & \textbf{16.98} \\

\bottomrule
\end{tabular}
}
\caption{Performance of various approaches on the LLaMA-13B backbone model at different target speedup ratios. 
The \textbf{bolded} scores correspond to the best performance under the same or comparable settings. "\#Target Activated Layer" means the average number of layers to be activated, and "\#Actual Activated layer" means the actual number of activated layers during the inference process of both tasks. 
"\sysnameshort" is short for "\sysname".
"RG-1", "RG-2" and "RG-1" are short for "ROUGE-1", "ROUGE-2" and "ROUGE-L", respectively.
$\downarrow$ indicates that smaller values are better, and $\uparrow$ indicates that larger values are better.
}
\label{tab:llama}
\end{center}
\end{table*}

\end{document}